\documentclass[sigconf]{acmart}

\usepackage{xspace}
\usepackage{graphicx}
\usepackage{bm}
\usepackage{marvosym}

\newcommand{\paratitle}[1]{\vspace{1.5ex}\noindent\textbf{#1}}
\newcommand{\ie}{\emph{i.e.,}\xspace}
\newcommand{\aka}{\emph{a.k.a.,}\xspace}
\newcommand{\eg}{\emph{e.g.,}\xspace}

\newcommand{\ignore}[1]{}

\copyrightyear{2022}
\acmYear{2022} 
\setcopyright{acmcopyright}
\acmConference[KDD '22]{Proceedings of the 28th ACM SIGKDD Conference on Knowledge Discovery and Data Mining}{August 14--18, 2022}{Washington, DC, USA}
\acmBooktitle{Proceedings of the 28th ACM SIGKDD Conference on Knowledge Discovery and Data Mining (KDD '22), August 14--18, 2022, Washington, DC, USA}
\acmPrice{15.00}
\acmDOI{10.1145/3534678.3539382}
\acmISBN{978-1-4503-9385-0/22/08}

\acmSubmissionID{rtfp1690}

\begin{document}

\title{Towards Unified Conversational Recommender Systems via Knowledge-Enhanced Prompt Learning}

\author{Xiaolei Wang}
\authornotemark[2]
\affiliation{%
  \institution{Gaoling School of Artificial Intelligence, Renmin University of China}
  \city{Beijing}
  \country{China}
}\email{wxl1999@foxmail.com}

\author{Kun Zhou}
\authornotemark[2]
\affiliation{%
  \institution{School of Information, Renmin University of China}
  \city{Beijing}
  \country{China}
}\email{francis_kun_zhou@163.com}\thanks{$^\dagger$Equal contribution.}

\author{Ji-Rong Wen}
\affiliation{%
  \institution{Gaoling School of Artificial Intelligence, Renmin University of China}
  \institution{Beijing Key Laboratory of Big Data Management and Analysis Methods}
  \city{Beijing}
  \country{China}
}\email{jrwen@ruc.edu.cn}

\author{Wayne Xin Zhao\textsuperscript{\Letter}}
\affiliation{
    \institution{Gaoling School of Artificial Intelligence, Renmin University of China}
    \institution{Beijing Key Laboratory of Big Data Management and Analysis Methods}
    \institution{Beijing Academy of Artificial Intelligence}
  \city{Beijing}
  \country{China}
}
\email{batmanfly@gmail.com}
\thanks{\textsuperscript{\Letter} Corresponding author.}


\begin{abstract}
Conversational recommender systems (CRS) aim to proactively elicit user preference and recommend high-quality items through natural language conversations.
Typically, a CRS consists of a recommendation module to predict preferred items for users and a conversation module to generate appropriate responses.
To develop an effective CRS, it is essential to seamlessly integrate the two modules.
Existing works either design semantic alignment strategies, or share knowledge resources and representations between the two modules.
However, these approaches still rely on different architectures or techniques to develop the two modules, making it difficult for effective module integration.

To address this problem, we propose a unified CRS model named \textbf{UniCRS} based on knowledge-enhanced prompt learning. 
Our approach unifies the recommendation and conversation subtasks into the prompt learning paradigm, and utilizes knowledge-enhanced prompts based on a fixed pre-trained language model (PLM) to fulfill both subtasks in a unified approach.
In the prompt design, we include fused knowledge representations, task-specific soft tokens, and the dialogue context, which can provide sufficient contextual information to adapt the PLM for the CRS task.
Besides, for the recommendation subtask, we also incorporate the generated response template as an important part of the prompt, to enhance the information interaction between the two subtasks.
Extensive experiments on two public CRS datasets have demonstrated the effectiveness of our approach.
Our code is publicly available at the link: \textcolor{blue}{\url{https://github.com/RUCAIBox/UniCRS}}.
\end{abstract}



\maketitle

\section{Introduction}
With the widespread of intelligent assistants, conversational recommender systems (CRSs) have become an emerging research topic, which provide the recommendation service to users through natural language conversations~\cite{christakopoulou2016towards,li2018towards}.
From the perspective of functions, CRSs should be able to fulfill two major subtasks, \emph{a recommendation subtask} that predicts items from a candidate set to users and \emph{a conversation subtask} that generates appropriate questions or responses.

To fulfill these two subtasks, existing methods~\cite{chen2019towards,zhou2020improving,li2021seamlessly} usually set up two separate modules for each subtask, namely the recommendation module and the conversation module.
Since the two subtasks are highly coupled, it has been widely recognized that a capable CRS should be able to seamlessly integrate these two modules~\cite{chen2019towards,zhou2020improving,li2021seamlessly,wang2022barcor}, in order to share useful features or knowledge between them.
One line of works incorporate shared knowledge resources (\eg knowledge graphs~\cite{chen2019towards} and reviews~\cite{lu2021revcore}) and their representations to enhance the semantic interaction.
Another line of works design special representation alignment strategies, such as pre-training tasks and regularization terms (\eg mutual information maximization~\cite{zhou2020improving} and contrastive learning~\cite{zhou2022c2}), to guarantee the semantic consistency of the two modules.

\begin{table}[t]
    \centering
    \caption{
        An illustrative case of the semantic inconsistency between the recommendation and conversation modules in existing CRS methods.
        The mentioned movies and entities are marked in italic blue and red, respectively.
        Compared with the baseline, the generated response of our model is more consistent with the predicted recommendation.
    }
    \label{tab:intro}
    \resizebox{\linewidth}{!}{%
        \begin{tabular}{p{0.2\linewidth}p{0.8\linewidth}}
            \toprule
            \textbf{USER:}     & Hello! I am looking for some movies.\\
            \textbf{HUMAN:}    & What kinds of movie do you like? I like \textcolor{red}{animated} movies such as \textcolor{blue}{\textit{Frozen (2013)}}.\\
            \textbf{USER:}     & I do not like \textcolor{red}{animated} films. I would love to see a movie like \textcolor{blue}{\textit{Pretty Woman (1990)}} starring \textcolor{red}{Julia Roberts}. Know any that are similar? \\
            \midrule
            \textbf{KGSF:} &
            \begin{tabular}[c]{@{}p{\linewidth}@{}}
                \textbf{Recommendation:} Frozen 2 (2019)\\
                \textbf{Response:} \textcolor{blue}{\textit{Pretty Woman (1990)}} is a great movie.
            \end{tabular}\\
            \midrule
            \textbf{OURS:}     &
            \begin{tabular}[c]{@{}p{\linewidth}@{}}
                \textbf{Recommendation:} My Best Friend's Wedding (1997)\\
                \textbf{Response:} Have you seen \textcolor{blue}{\textit{My Best Friend's Wedding (1997)}}? \textcolor{red}{Julia Roberts} also stars in it.
            \end{tabular}\\
            \midrule
            \textbf{HUMAN:}    & \textcolor{blue}{\textit{Pretty Woman (1990)}} was a good one. If you are in it for \textcolor{red}{Julia Roberts} you can try \textcolor{blue}{\textit{Runaway Bride (1999)}}.\\
            \bottomrule
        \end{tabular}%
    }
\end{table}

Despite the progress of existing CRS methods, the fundamental issue of semantic inconsistency between the recommendation and conversation modules has not been well addressed.
Figure~\ref{tab:intro} shows an inconsistent case of the prediction from a representative CRS model, KGSF~\cite{zhou2020improving}, which utilizes mutual information maximization to align the semantic representations.
Although the recommendation module predicts the movie \textit{``Frozen 2 (2019)''}, the conversation module seems to be unaware of such a recommendation result and generates a mismatched response that contains another movie \textit{``Pretty Woman (1990)''}.
Even if we can utilize heuristic constraints to enforce the generation of the recommended movie, it cannot fundamentally resolve the semantic inconsistency of the two modules.
In essence, such a problem is caused by two major issues in existing methods. 
First, most of these methods develop the two modules with different architectures or techniques. 
Even with some shared knowledge or components, it is still difficult to effectively associate the two modules seamlessly. 
Second, results from one module cannot be perceived and utilized by the other. 
For example, there is no way to leverage the generated response when predicting the recommendation results in KGSF~\cite{zhou2020improving}.
To summarize, the root of semantic inconsistency is the different architecture designs and working mechanisms of the two modules. 

To address the above issues, we aim to develop a more effective CRS that implements both the recommendation and conversation modules in a unified manner. 
Our approach is inspired by the great success of pre-trained language models (PLMs)~\cite{jiang2020can,gao2021making,brown2020language}, which have been shown effective as a general solution to a variety of tasks even in very different settings. 
In particular, the recently proposed paradigm \emph{prompt learning}~\cite{brown2020language,gao2021making,wang2021finetuning} further unifies the use of PLMs on different tasks in a simple yet flexible manner. 
Generally speaking, prompt learning augments or extends the original input of PLMs by prepending explicit or latent tokens, which might contain demonstrations, instructions, or learnable embeddings.
Such a paradigm can unify different task formats or data forms to a large extent.
For CRSs, since the two subtasks aim to fulfill specific goals based on the same conversational semantics, it is feasible to develop a unified CRS approach based on prompt learning.

To this end, in this paper, we propose a novel unified CRS model based on knowledge-enhanced prompt learning, namely \textbf{UniCRS}. 
For the base PLM, we utilize DialoGPT~\cite{zhang2020dialogpt} since it has been pre-trained on a large-scale dialogue corpus. 
In our approach, the base PLM is \emph{fixed} in solving the two subtasks, without fine-tuning or continual pre-training. 
To better inject the task knowledge into the base PLM, we first design a semantic fusion module that can capture the semantic association between words from dialogue texts and entities from knowledge graphs~(KGs).
The major technical contribution of our approach lies in that we formulate the two subtasks in the form of prompt learning, and design specific prompts for each subtask.
In our prompt design, we include the dialogue context (\emph{specific tokens}), task-specific soft tokens (\emph{latent vectors}), and fused knowledge representations (\emph{latent vectors}), which can provide sufficient semantic information about the dialogue context, task instructions, and background knowledge. 
Moreover, for recommendation, we incorporate the generated response templates from the conversation module into the prompt, which can further enhance the information interaction between the two subtasks.

To validate the effectiveness of our approach, we conduct experiments on two public CRS datasets. 
Experimental results show that our UniCRS outperforms several competitive methods on both the recommendation and conversation subtasks, especially when training data is limited. Our main contributions are summarized as:

(1) To the best of our knowledge, it is the first time that a unified CRS has been developed in a general prompt learning way.

(2) Our approach formulates the subtasks of CRS into a unified form of prompt learning, and designs task-specific prompts with corresponding optimization methods.

(3) Extensive experiments on two public CRS datasets have demonstrated the effectiveness of our approach in both the recommendation and conversation tasks.

\section{Related Work}
Our work is related to the following two research directions, namely conversational recommendation and prompt learning.

\subsection{Conversational Recommendation}
With the rapid development of dialogue systems~\cite{chen2017survey,zhang2020dialogpt}, conversational recommender systems~(CRSs) have emerged as a research topic, which aim to provide accurate recommendations through conversational interactions with users~\cite{christakopoulou2016towards,sun2018conversational,gao2021advances}.
A major category of CRS studies rely on pre-defined actions (\eg intent slots or item attributes) to interact with users~\cite{christakopoulou2016towards,sun2018conversational,zhou2020leveraging}.
They focus on accomplishing the recommendation task within as few turns as possible.
They adopt the multi-armed bandit model~\cite{christakopoulou2016towards,xie2021comparison} or reinforcement learning~\cite{sun2018conversational} to find the optimal interaction strategy.
However, methods that belong to this category mostly rely on pre-defined actions and templates to generate responses, which largely limit their usage in various scenarios.
Another category of CRS studies aim to generate both accurate recommendations and human-like responses~\cite{li2018towards,zhou2020towards,hayati2020inspired}.
To achieve this, these works usually devise a recommendation module and a conversation module to implement the two functions, respectively.
However, such a design raises the issue of semantic inconsistency, and it is essential to seamlessly integrate the two modules as a system.
Existing works mostly either share the knowledge resources and their representations~\cite{chen2019towards,lu2021revcore}, or design semantic alignment pre-training tasks~\cite{zhou2020improving} and regularization terms~\cite{zhou2022c2}.
However, it is still difficult for the effective integration of the two modules due to their different architectures or techniques. 
For example, it has been pointed out that the generated responses from the conversation module do not always match the predicted items from the recommendation module~\cite{liang2021learning}.
Our work follows the latter category and adopts prompt learning based on pre-trained language models (PLM) to unify the recommendation and conversation subtasks.
In this way, the two subtasks can be formulated in a unified manner with elaborately designed prompts.

\subsection{Prompt Learning}
Recent years have witnessed the remarkable performance of PLMs on a variety of tasks~\cite{devlin2019bert,lewis2020bart}.
Most of PLMs are pre-trained with the objective of language modeling but are fine-tuned on downstream tasks with quite different objectives.
To overcome the gap between pre-training and fine-tuning, prompt learning (\aka prompt-tuning) has been proposed~\cite{liu2021pre,gu2021ppt}, which relies on carefully designed prompts to reformulate the downstream tasks as the pre-training task.
Early works mostly incorporate manually crafted discrete prompts to guide the PLM~\cite{brown2020language,raffel2020exploring}.
Recently, a surge of works focus on automatically optimizing discrete prompts for specific tasks~\cite{gao2021making,jiang2020can} and achieving comparable performance with manual prompts.
However, these methods still rely on generative models or complex rules to control the quality of prompts.
In contrast, some works propose to use learnable continuous prompts that can be directly optimized~\cite{li2021prefix,lester2021power}.
On top of this, several works devise prompt pre-training tasks~\cite{gu2021ppt} or knowledgeable prompts~\cite{hu2021knowledgeable} to improve the quality of the continuous prompts.
In this work, we reformulate both the recommendation and conversation subtasks as the pre-training task of a PLM by prompt learning.
In addition, to provide the PLM with task-related knowledge of CRS, we enhance the prompts with the information from an external KG and perform semantic fusion for prompt learning.

\section{Problem Statement}
Conversational recommender systems (CRSs) aim to conduct item recommendation through multi-turn natural language conversations.
At each turn, the system either makes recommendations or asks clarification questions, based on the currently learned user preference.
Such a process ends until the user accepts the recommended items or leaves.
Typically, a CRS consists of two modules, \ie the recommender module and the conversation module, which are responsible for the recommendation and the response generation tasks, respectively.
These two modules should be seamlessly integrated to generate consistent results, in order to fulfill the conversational recommendation task.

Formally, let $u$ denote a user, $i$ denote an item from the item set $\mathcal{I}$, and $w$ denote a word from the vocabulary $\mathcal{V}$.
A conversation is denoted as $C=\{s_t\}_{t=1}^n$, where $s_t$ denotes the utterance at the $t$-th turn and each utterance $s_t=\{w_j\}_{j=1}^m$ consists of a sequence of words from the vocabulary $\mathcal{V}$.

With the above definitions, the task of conversational recommendation is defined as follows.
At the $t$-th turn, given the dialogue history $C=\{s_j\}_{j=1}^{t-1}$ and the item set $\mathcal{I}$, the system should (1) select a set of candidate items $\mathcal{I}_t$ from the entire item set $\mathcal{I}$ to recommend, and (2) generate the response $R=s_{t}$ that includes the items in $\mathcal{I}_t$.
Note that $\mathcal{I}_t$ might be empty, when there is no need for recommendation.

\section{Approach}

\begin{figure*}
	\includegraphics[width=\textwidth]{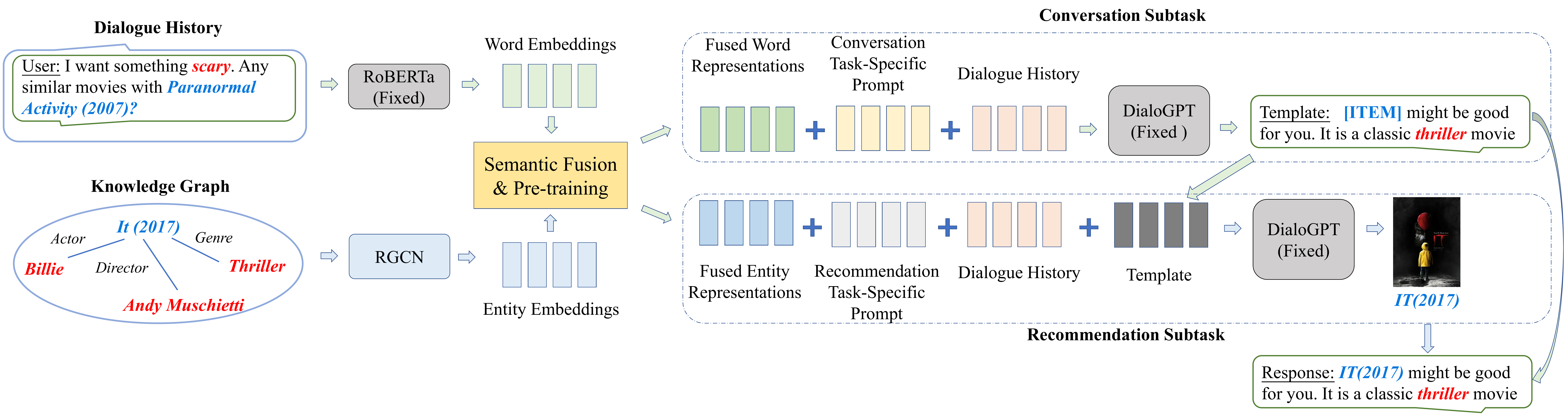}
	\centering
	\caption{
	    The overview of the proposed framework UniCRS.
	    Blocks in grey indicate that their parameters are frozen, while other parameters are tunable.
	    We first perform pre-training to fuse semantics from both words and entities, then prompt the PLM to generate the response template and use the template as part of the prompt for recommendation. 
	    Finally, the recommended items are filled into the template as a complete response.
    }
	\label{fig:approach}
\end{figure*}

In this section, we present a unified CRS approach with knowledge-enhanced prompt learning based on a PLM, namely \textbf{UniCRS}. 
We first give an overview of our approach, then discuss how to fuse semantics from words and entities as part of the prompts, and finally present the knowledge-enhanced prompting approach to the CRS task. 
The overall architecture of our proposed model is presented in Figure~\ref{fig:approach}.

\subsection{Overview of the Approach}
Previous studies on CRS~\cite{li2018towards,chen2019towards,zhou2020improving} usually develop specific modules for the recommendation and conversation subtasks respectively, and they need to connect the two modules in order to fulfill the task goal of CRS.
Different from existing CRS methods, we aim to develop a unified approach with prompt learning based on PLM.

\paratitle{The Base PLM}.
In our approach, we take DialoGPT~\cite{zhang2020dialogpt} as our base PLM.
DialoGPT adopts a Transformer-based autoregressive architecture and is pre-trained on a large-scale dialogue corpus extracted from Reddit.
It has been shown that DialoGPT can generate coherent and informative responses, making it a suitable base model for the CRS task~\cite{wang2021finetuning,liang2021learning}. 
Let $f(\cdot \mid \Theta_{plm})$ denote the base PLM parameterized by $\Theta_{plm}$, taking a token sequence as input and producing contextualized representations for each token.
Unless otherwise specified, we will use the representation of the last token from DialoGPT for subsequent prediction or generation tasks.

\paratitle{A Unified Prompt-Based Approach to CRS}. 
Given the dialogue history $\{s_j\}_{j=1}^{t-1}$ at the $t$-th turn, we concatenate each utterance into a text sequence $C=\{w_k\}_{k=1}^{n_W}$.
The basic idea is to encode the dialogue history $C$, obtain its contextualized representations, and solve the recommendation and conversation subtasks via \emph{generation} (\ie generating either \emph{the recommended items} or \emph{the response utterance}), with the base PLM.
In this way, the two subtasks can be fulfilled in a unified approach.
However, since the base PLM is fixed, it is difficult to achieve satisfactory performance compared with fine-tuning due to lack of task adaptation.
Therefore, we adopt the prompting approach~\cite{gao2021making,gu2021ppt}, where the original dialogue history is prepended with elaborately designed or learned \emph{prompt tokens}, denoted by $\{p_k\}_{k=1}^{n_P}$ ($n_P$ is the number of prompt tokens).
In practice, prompt tokens can be either explicit tokens or latent vectors.
It has been shown that prompting is an effective paradigm to leverage the knowledge of PLMs to solve various tasks without fine-tuning~\cite{brown2020language,gao2021making}.

\paratitle{Prompt-augmented Dialogue Context}.
By incorporating the prompts, the original dialogue history $C$ can be extended to a longer sequence (called \emph{context sequence}), denoted as $\widetilde{C}$:
\begin{equation}
    \widetilde{C} \rightarrow \underbrace{p_1,\dots, p_{n_P},}_{\text{prompt tokens}} \underbrace{w_{1}\cdots w_{n_W}}_{\text{word tokens}}.
\end{equation}

As before, we utilize the base PLM to obtain contextualized representations of the context sequence for solving the recommendation and conversation subtasks.
In order to better adapt to the task characteristics, we can construct and learn different prompts, and obtain corresponding context sequences denoted as $\widetilde{C}_{rec}$ for recommendation and $\widetilde{C}_{con}$ for conversation.

To implement such a unified approach, we identify two major problems to solve: (1) how to fuse conversational semantics and related knowledge semantics in order to adapt the base PLM for CRS (Section~\ref{sec-sf}), and (2) how to design and learn suitable prompts for the recommendation and conversation subtasks (Section~\ref{sec-4.3}).  
In what follows, we will introduce the two parts in detail. 

\subsection{Semantic Fusion for Prompt Learning}
\label{sec-sf}
Since DialoGPT is pre-trained on a general dialogue corpus, it lacks the specific capacity for the CRS task and cannot be directly used.
Following previous studies~\cite{chen2019towards,zhou2020improving}, we incorporate KGs as the task-specific knowledge resources, since it involves useful knowledge about entities and items mentioned in the dialogue.
However, it has been found that there is a large semantic gap between the semantic spaces of dialogues and KGs~\cite{zhou2020improving,zhou2022c2}. 
We need to first fuse the two semantic spaces for effective knowledge alignment and enrichment.
Specially, the purpose of this step is to fuse the token and entity embeddings from different encoders.

\paratitle{Encoding Word Tokens and KG Entities.} 
Given a dialogue history $C$, we first separately encode the dialogue words and KG entities that appear in $C$ into word embeddings and entity embeddings.  
To complement our base PLM DialoGPT~(a unidirectional decoder), we employ another fixed PLM RoBERTa~\cite{liu2019roberta}~(a bi-directional encoder) to derive the word embeddings.
The contextualized token representations derived from the fixed encoder RoBERTa are concatenated into a word embedding matrix, \ie $\mathbf{T}=[\bm{h}_1^T; \dots; \bm{h}_{n_W}^T]$. 
For entity embeddings, following previous works~\cite{chen2019towards,zhou2020improving}, we first perform entity linking based on an external KG DBpedia~\cite{auer2007dbpedia}, and then obtain the corresponding entity embeddings via a relational graph neural networks~(RGCN)~\cite{schlichtkrull2018modeling}, which can model the relational semantics through information propagation and aggregation over the KG.
Similarly, the derived entity embedding matrix is denoted as $\mathbf{E}=[\bm{h}_1^E; \dots; \bm{h}_{n_E}^E]$, where $n_E$ is the number of mentioned entities in the dialogue history.

\paratitle{Word-Entity Semantic Fusion.}
In order to bridge the semantic gap between words and entities, we use a cross interaction mechanism to associate the two kinds of semantic representations via a bilinear transformation:
\begin{align}
    \mathbf{A}              & =\mathbf{T}^{\top}\mathbf{W} \mathbf{E},    \label{eq-A} \\ 
    \widetilde{\mathbf{T}}  & =\mathbf{T}+\mathbf{E}\mathbf{A},             \label{eq-T}  \\ 
    \widetilde{\mathbf{E}}  & =\mathbf{E}+\mathbf{T}\mathbf{A}^{\top},      \label{eq-E}
\end{align}
where $\mathbf{A}$ is the affinity matrix between the two representations, $\mathbf{W}$ is the transformation matrix, $\widetilde{\mathbf{T}}$ is the fused word representations, and $\widetilde{\mathbf{E}}$ is the fused entity representations.
Here we use the bilinear transformation between $\mathbf{T}$ and $ \mathbf{E}$ for simplicity, and leave the further exploration of complex interaction mechanisms for future work.

\paratitle{Pre-training the Fusion Module}.
After semantic fusion, we can establish the semantic association between words and entities. 
However, such a module involves additional learnable parameters, denoted as $\Theta_{fuse}$. 
To better optimize the parameters of the fusion module, we propose a prompt-based pre-training approach that leverages the self-supervision signals from the dialogues.
Specifically, we prepend the fused entity representations $\widetilde{\mathbf{E}}$ (Eq.~\ref{eq-E}) and append the response to the dialogue context, namely $\widetilde{C}_{pre}=[\widetilde{\mathbf{E}}; C; R]$, where we use the \underline{\emph{bold font}} to denote the \emph{latent vectors} ($\widetilde{\mathbf{E}}$) and the \underline{\emph{plain font}} to denote the \emph{explicit tokens} ($C, R$).
For this pre-training task, we simply utilize the prompt-augmented context sequence $\widetilde{C}_{pre}$ to predict the entities appearing in the response.
The prediction probability of the entity $e$ is formulated as:
\begin{align}
    \label{eq:rec-ent}
    \text{Pr}(e \mid \widetilde{C}_{pre}) & =\text{Softmax}(\bm{h}_u \cdot\bm{h}_e),
\end{align}
where $\bm{h}_u= \text{Pooling}[ f(\widetilde{C}_{pre} \mid \Theta_{plm}; \Theta_{fuse}) ]$ is the learned representation of the context by pooling the contextualized representations of all the tokens in $\widetilde{C}_{pre}$, and $\bm{h}_e$ is the fused entity representation for the entity $e$.
Note that only the parameters of the fusion module $\Theta_{fuse}$ are required to optimize, while the parameters of the base PLM $\Theta_{plm}$ are fixed.
We adopt the cross-entropy loss for the pre-training task.

After semantic fusion, we obtain the fused knowledge representations for words and entities from the dialogue history, namely $\widetilde{\mathbf{T}}$ (Eq.~\ref{eq-T}) and $\widetilde{\mathbf{E}}$ (Eq.~\ref{eq-E}), respectively. 
These representations are subsequently used as part of prompts, as shown in Section~\ref{sec-4.3}.

\subsection{Subtask-specific Prompt Design}
\label{sec-4.3}
Though the base PLM is fixed without fine-tuning, we can design specific prompts to adapt it to different subtasks of CRS. 
For each subtask (either \emph{recommendation} or \emph{conversation}), the major design of prompting consists of three parts, namely the dialogue history, subtask-specific soft tokens, and fused knowledge representations.
For recommendation, we further incorporate the generated response templates as additional prompt tokens. 
Next, we describe the specific prompting designs for the two subtasks in detail.

\subsubsection{Prompt for Response Generation}
The subtask of response generation aims to generate informative utterances in order to clarify user preferences or reply to users' utterances.
The prompting design mainly enhances the textual semantics for better dialogue understanding and response generation.

\paratitle{The Prompt Design}. 
The prompt for response generation consists of the original dialogue history (in the form of \emph{word tokens}  $C$), generation-specific soft tokens (in the form of \emph{latent vectors} $\mathbf{P}_{gen}$) and fused textual context (in the form of \emph{latent vectors} $\widetilde{\mathbf{T}}$), which is formally denoted as:
\begin{equation}
\widetilde{C}_{gen} \rightarrow [\text{~}\widetilde{\mathbf{T}}; \text{~~~~~}\mathbf{P}_{gen}; \text{~~~~~} C\text{~} ],
\label{eq-gen-prompt}
\end{equation}
where we use the \emph{bold} and \emph{plain} fonts to denote soft and hard token sequences, respectively.
In this design, the subtask-specific prompts $\mathbf{P}_{gen}$ instruct the PLM by the signal from the generation task, the KG-enhanced textual representations $\widetilde{\mathbf{T}}$ (Eq.~\ref{eq-T}), and the original dialogue history $C$.

\paratitle{Prompt Learning}. 
In the above prompting design, the only tunable parameters are the fused textual representations $\widetilde{\mathbf{T}}$ that have been pre-trained, and generation-specific soft tokens $\mathbf{P}_{gen}$.
They are denoted as $\Theta_{gen}$. 
We use the prompt-augmented context $\widetilde{C}_{gen}$ to derive the prediction loss for learning $\Theta_{gen}$, which is formally given as:
\begin{align}
    L_{gen}(\Theta_{gen})   &= -\frac{1}{N}\sum_{j=1}^N\log \text{Pr}(R_j \mid \widetilde{C}_{gen}^{(j)} ; \Theta_{gen})    \notag \\
                            &= -\frac{1}{N}\sum_{i=1}^N\sum_{j=1}^{l_i}\log \text{Pr}(w_{i,j} \mid \widetilde{C}_{gen}^{(j)} ; \Theta_{gen} ; w_{<j}),
 \label{eq:conv-loss-p}
\end{align}
where $N$ is the number of training instances (a pair of the context and target utterances), and $l_i$ is the length of the $i$-th target utterance, and $w_{<j}$ denotes the words proceeding the $j$-th position.

\paratitle{Response Template Generation}. 
Besides sharing the base PLM, we find that it is also important to share intermediate results of different subtasks to achieve more consistent final results.
For example, given the generated response of the conversation task, the PLM might be able to predict more relevant recommendations according to such extra contextual information.
Based on this intuition, we propose to include response templates as part of the prompt for the recommendation subtask.
Specifically, we add a special token $\texttt{[ITEM]}$ into the vocabulary $\mathcal{V}$ of the base PLM and replace all the items that appear in the response with the $\texttt{[ITEM]}$ token.
At each time step, the PLM generates either the special token $\texttt{[ITEM]}$ or a general token from the original vocabulary.
All the slots will be filled after the recommended items are generated. 

\subsubsection{Prompt for Item Recommendation}
The subtask of recommendation aims to predict items that a user might be interested in.
The prompting design mainly enhances the user preference semantics, in order to predict more satisfactory recommendations.

\paratitle{The Prompt Design}. 
The item recommendation prompts consist of the original dialogue history $C$ (in the form of \emph{word tokens}), recommendation-specific soft tokens $\mathbf{P}_{rec}$ (in the form of \emph{latent vectors}), fused entity context $\widetilde{\mathbf{E}}$ (in the form of \emph{latent vectors}), and the response template $S$ (in the form of \emph{word tokens}), formally described as:
\begin{equation}
\widetilde{C}_{rec} \rightarrow [\text{~}\widetilde{\mathbf{E}}; \text{~~~~~}\mathbf{P}_{rec}; \text{~~~~~} C; \text{~~~~~} S \text{~} ],
\label{eq-rec-prompt}
\end{equation}
where the subtask-specific prompts $\mathbf{P}_{rec}$ instruct the PLM by the signal from the recommendation task, the KG-enhanced entity representations $\widetilde{\mathbf{E}}$ (Eq.~\ref{eq-E}), the original dialogue history $C$, and the response template $S$.

A key difference between the prompts of the two subtasks is that we utilize entity representations for \emph{recommendation}, and word representations for \emph{generation}. This is because their prediction targets are items and sentences, respectively. 
Besides, we have a special design for recommendation, where we include the response template as part of the prompts.
This can enhance the subtask connections and alleviate the risk of semantic inconsistency.

\paratitle{Prompt Learning}. 
In the above prompting design, the only tunable parameters are the fused entity representations $\widetilde{\mathbf{E}}$ that have been pre-trained, and recommendation-specific soft tokens $\mathbf{P}_{gen}$.
They are denoted as $\Theta_{rec}$. 
We utilize the prompt-augmented context $\widetilde{C}_{rec}$ to derive the prediction loss for learning $\Theta_{rec}$, which is formally given as:
\begin{equation}
    \label{eq:rec-loss-p}
    \small
    L_{rec}(\Theta_{rec})=-\sum_{j=1}^N\sum_{i=1}^M\big[y_{j,i}\cdot\log\text{Pr}_{j}(i)+(1-y_{j,i})\cdot\log(1-\text{Pr}_{j}(i))\big],
\end{equation}
where $N$ is the number of training instances (a pair of the context and a target item), $M$ is the total number of items, $y_{j,i}$ denotes a binary ground-truth label which is equal to 1 when item $i$ is the correct label for the $j$-th training instance, and $\text{Pr}_{j}(i)$ is an abbreviation of $\text{Pr}(i \mid \widetilde{C}_{rec}^{(j)} ; \Theta_{rec})$, which is computed following a similar way in Eq.~\ref{eq:rec-ent} by first pooling contextualized representations and then computing the softmax score.

\subsection{Parameter Learning}
The parameters of our model consist of four groups, namely the base PLM, the semantic fusion module, and the subtask-specific soft tokens for recommendation and conversation.
They are denoted as $\Theta_{plm}$, $\Theta_{fuse}$, $\Theta_{rec}$ and $\Theta_{gen}$, respectively.

During the overall training process, the parameters of the base PLM $\Theta_{plm}$ are always fixed, and we only optimize the rest parameters.
First, we pre-train the parameters of the semantic fusion module $\Theta_{fuse}$. 
Given the dialogue history and KG, we encode the dialogue tokens with a fixed text encoder RoBERTa and the KG entities with a learnable graph encoder RGCN.
Then, we perform semantic fusion to obtain the fused word representations $\widetilde{\mathbf{T}}$ using Eq.~\ref{eq-T} and entity representations $\widetilde{\mathbf{E}}$ using Eq.~\ref{eq-E}.
After that, we optimize $\Theta_{fuse}$ based on the self-supervised entity prediction task.
Next, we randomly initialize the parameters of the subtask-specific soft tokens $\Theta_{rec}$ and $\Theta_{gen}$, and compose the response generation prompts using Eq.~\ref{eq-gen-prompt}.
We utilize the supervised signal from the conversation task to learn $\Theta_{gen}$ using Eq.~\ref{eq:conv-loss-p} and generate the response template.
Finally, we compose the item recommendation prompts using Eq.~\ref{eq-rec-prompt} and leverage the supervised signal from the recommendation task to learn $\Theta_{rec}$ using Eq.~\ref{eq:rec-loss-p}.
\section{Experiment}

In this section, we first set up the experiments, and then report the results and give detailed analysis.

\subsection{Experimental Setup}

\begin{table}[t]
    \centering
    \caption{Statistics of the datasets after preprocessing.}
    \small
    \label{tab:datasets}
    \begin{tabular}{crrr}
        \toprule
        \textbf{Dataset} & \textbf{\#Dialogs} & \textbf{\#Utterances} & \textbf{\#Items} \\
        \midrule
        INSPIRED         & 1,001              & 35,811                & 1,783            \\
        ReDial           & 10,006             & 182,150               & 51,699           \\
        \bottomrule
    \end{tabular}
\end{table}

\paratitle{Datasets.}
To evaluate the performance of our model, we conduct experiments on the \textsc{ReDial}~\cite{li2018towards} and \textsc{INSPIRED}~\cite{hayati2020inspired} datasets.
The \textsc{ReDial} dataset is an English CRS dataset about movie recommendations, and is constructed through crowd-sourcing workers on Amazon Mechanical Turk (AMT).
Similar to \textsc{ReDial}, the \textsc{INSPIRED} dataset is also an English CRS dataset about movie recommendations, but with a smaller size.
These two datasets are widely used for evaluating CRS models. 
The statistics of both datasets are summarized in Table~\ref{tab:datasets}.

\paratitle{Baselines.}
For CRS, we consider two major subtasks for evaluation, namely recommendation and conversation.
For comparison, we select several representative methods (including both CRS models and adapted PLMs) tailored to each subtask.

\textbullet~\underline{\textbf{ReDial}}~\cite{li2018towards}:
It is proposed along with the \textsc{ReDial} dataset, which incorporates a conversation module based on HRED~\cite{subramanian2018learning} and a recommendation module based on auto-encoder~\cite{sedhain2015autorec}.

\textbullet~\underline{\textbf{KBRD}}~\cite{chen2019towards}:
It utilizes an external KG to enhance the semantics of entities mentioned in the dialogue history, and adopts a self-attention based recommendation module and a Transformer-based conversation module.

\textbullet~\underline{\textbf{KGSF}}~\cite{zhou2020improving}:
It incorporates two KGs to enhance the semantic representations of words and entities, and utilizes the Mutual Information Maximization method to align the semantic spaces of the two KGs.

\textbullet~\underline{\textbf{GPT-2}}~\cite{radford2019language}:
It is an auto-regressive PLM. 
We concatenate the historical utterances of a conversation as the input, and take the generated text as the response and the representation of the last token for recommendation.

\textbullet~\underline{\textbf{DialoGPT}}~\cite{zhang2020dialogpt}:
It is an auto-regressive model pre-trained on a large-scale dialogue corpus. 
Similar to GPT-2, we also adopt the generated text and the last token representation for the conversation and recommendation tasks, respectively.

\textbullet~\underline{\textbf{BERT}}~\cite{devlin2019bert}:
It is pre-trained via the masked language model task on a large-scale general corpus. 
We utilize the representation of the $[CLS]$ token for recommendation.

\textbullet~\underline{\textbf{BART}}~\cite{lewis2020bart}:
It is a seq2seq model pre-trained with the denoising auto-encoding task on a large-scale general corpus.
We also adopt the generated text and the last token representation for the conversation and recommendation tasks, respectively.

Among these baselines, ReDial~\cite{li2018towards}, KBRD~\cite{chen2019towards} and KGSF~\cite{zhou2020improving} are conversational recommendation methods, where the latter two incorporate external knowledge graphs;
BERT~\cite{devlin2019bert}, GPT-2~\cite{radford2019language}, BART~\cite{lewis2020bart}, and DialoGPT~\cite{zhang2020dialogpt} are pre-trained language models, where BERT, GPT-2 and BART are pre-trained on a general corpus, and DialoGPT is pre-trained on a dialogue corpus.

\paratitle{Evaluation Metrics.}
Following previous CRS works~\cite{li2018towards,zhou2020improving}, we adopt different metrics to evaluate the recommendation and conversation task separately.
For the recommendation task, following~\cite{chen2019towards,zhou2020improving}, we use Recall@$k$ ($k$=1,10,50) for evaluation.
For the conversation task, following~\cite{chen2019towards,zhou2020improving}, we adopt Distinct-$n$ ($n$=2,3,4) at the word level to evaluate the diversity of the generated responses.
Besides, following KGSF~\cite{zhou2020improving}, we invite three annotators to score the generated responses of our model and baselines from two aspects, namely \emph{Fluency} and \emph{Informativeness}. The range of scores is 0 to 2.
For all the above metrics, we calculate and report the average scores on all test examples.

\paratitle{Implementation Details.}
We select the DialoGPT-small model as the base PLM, which is pre-trained on 147M dialogues collected from Reddit.
It consists of 12 transformer layers, and the dimension of its embeddings is 768.
We freeze all its parameters during the overall training process.
To be consistent with DialoGPT-small, the hidden size of our designed prompts is also set to 768.
In the semantic fusion module, we utilize a fixed RoBERTa-base model for encoding the input tokens, and set the layer number of R-GCN to 1 following KGSF~\cite{zhou2020improving}.
Besides, we set the length of soft prompt tokens to 10 for the recommendation task and 50 for the conversation task according to our parameter tuning results.
We use AdamW~\cite{loshchilov2018decoupled} with the default parameter setting to optimize the tunable parameters in our approach.
The batch size is set to 64 for the recommendation subtask and 8 for the conversation subtask, and the learning rate is 0.0005 for prompt pre-training and 0.0001 for the two subtasks.
We implement all baseline models using the open-source toolkit CRSLab~\cite{zhou2021crslab}~\footnote{https://github.com/RUCAIBox/CRSLab}, which contains comprehensive conversational recommendation models and benchmark datasets.

\subsection{Evaluation on Recommendation Task}
In this part, we conduct experiments to evaluate the effectiveness of our model on the recommendation task.

\begin{table}[t]
    \centering
    \caption{
        Results on the recommendation task.
        Numbers marked with * indicate that the improvement is statistically significant compared with the best baseline (t-test with p-value < 0.05).
    }
    \label{tab:rec-table}
    \resizebox{\linewidth}{!}{%
        \begin{tabular}{lcccccc}
            \toprule
            Datasets & \multicolumn{3}{c}{ReDial} & \multicolumn{3}{c}{INSPIRED}                                                                         \\
            \midrule
            Models   & R@1                        & R@10                         & R@50            & R@1             & R@10            & R@50            \\
            \midrule
            ReDial   & 0.023                      & 0.129                        & 0.287           & 0.003           & 0.117           & 0.285           \\
            KBRD     & 0.033                      & 0.175                        & 0.343           & 0.058           & 0.146           & 0.207           \\
            KGSF     & 0.035                      & 0.177                        & 0.362           & 0.058           & 0.165           & 0.256           \\
            \midrule
            GPT-2    & 0.023                      & 0.147                        & 0.327           & 0.034           & 0.112           & 0.278           \\
            DialoGPT & 0.030                      & 0.173                        & 0.361           & 0.024           & 0.125           & 0.247           \\
            BERT     & 0.030                      & 0.156                        & 0.357           & 0.044           & 0.179           & 0.328           \\
            BART     & 0.034                      & 0.174                        & 0.377           & 0.037           & 0.132           & 0.247           \\
            \midrule
            UniCRS   & \textbf{0.051}*            & \textbf{0.224}*              & \textbf{0.428}* & \textbf{0.094}* & \textbf{0.250}* & \textbf{0.410}* \\
            \bottomrule
        \end{tabular}%
    }
\end{table}

\begin{figure}[t]
    \centering
    \includegraphics[width=0.49\linewidth]{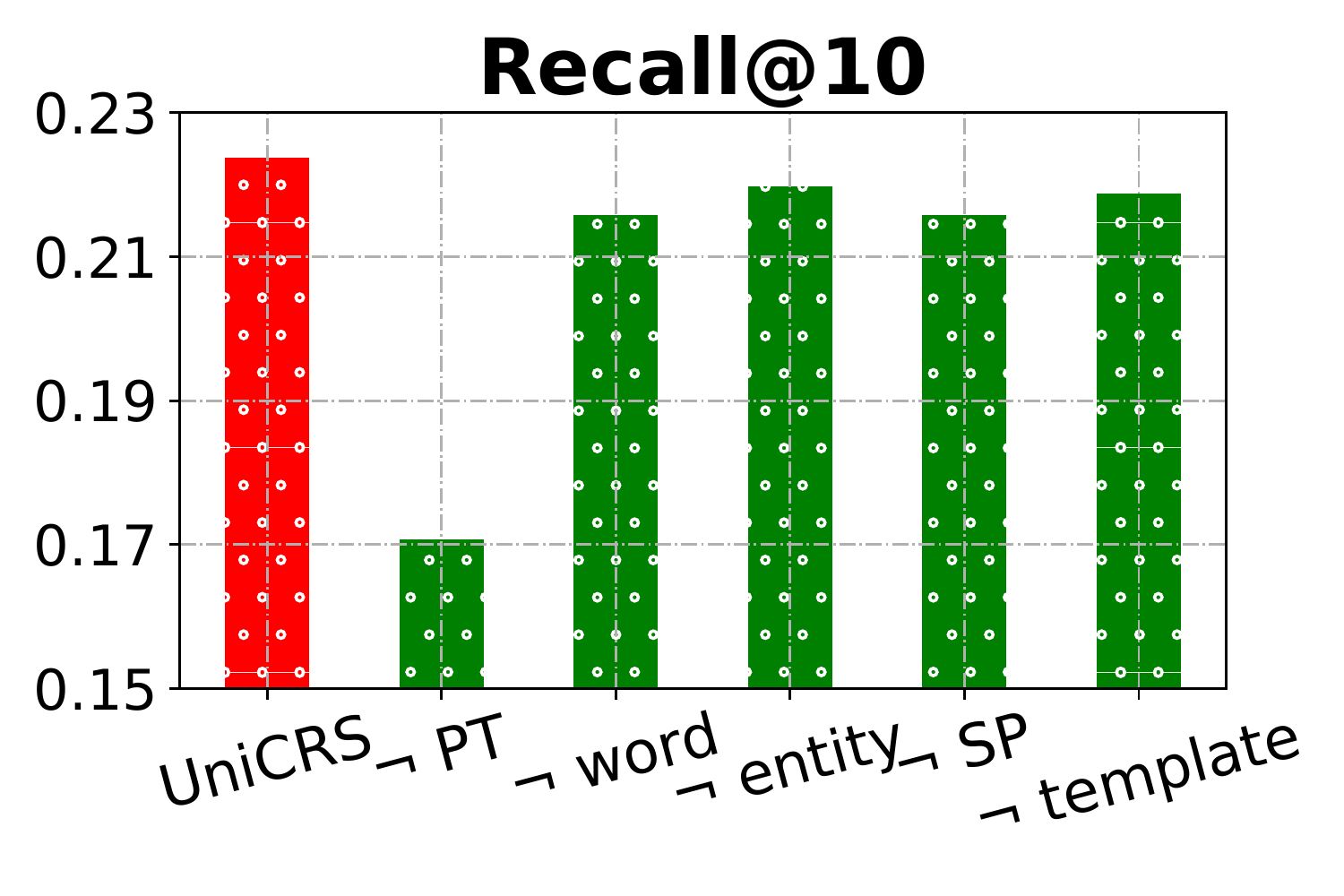}
    \includegraphics[width=0.49\linewidth]{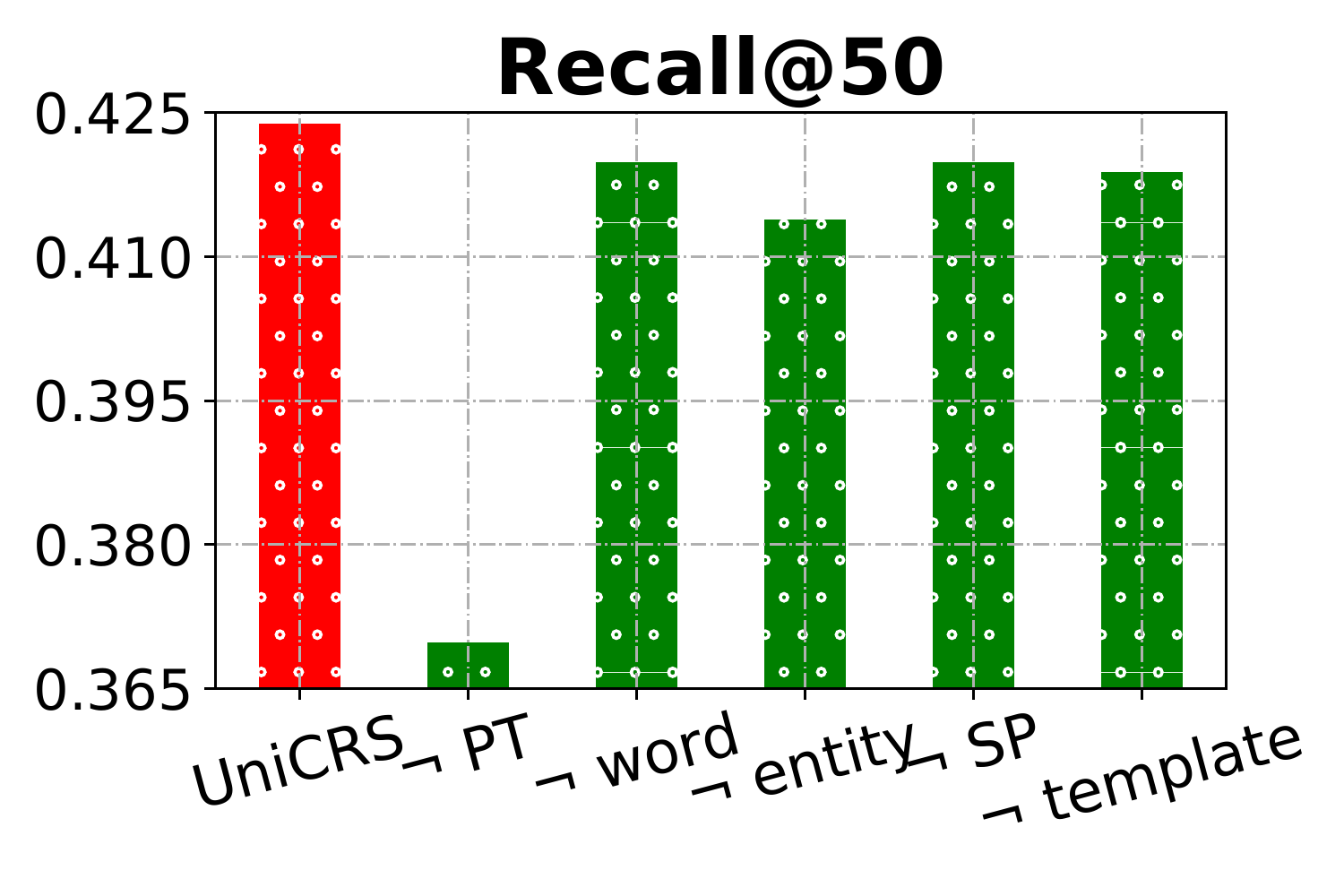}
    \caption{
        Ablation study on the \textsc{ReDial} dataset about the recommendation task. 
        PT denotes the pre-training task of semantic fusion. 
        Word and entity refer to two kinds of data signals in the fusion module. 
        SP and template refer to task-specific soft tokens and response templates, respectively.
    }
    \label{fig:ablation-rec}
\end{figure}

\paratitle{Automatic Evaluation.}
Table~\ref{tab:rec-table} shows the performance of different methods on the recommendation task.
For the three CRS methods, the performance order is consistent cross all datasets, \ie \emph{KGSF > KBRD > ReDial}.
KGSF and KBRD both incorporate external KGs into their recommendation modules, which can enrich the semantics of entities mentioned in the dialogue history to better capture user intents and preferences.
Besides, KGSF also adopts the mutual information maximization method to further improve the entity representations.
For the four pre-trained models, we can see that BERT and BART perform better than GPT-2 and DialoGPT. The reason might be that GPT-2 and DialoGPT are based on unidirectional Transformer architecture, which limits their capacity of dialogue understanding.
Furthermore, we can see that BART achieves comparable performance and even outperforms BERT on the \textsc{ReDial} dataset. It indicates that BART can also understand the dialogue semantics well for the recommendation task.

Finally, we can see that our model outperforms all the baselines by a large margin. 
We utilize specially designed prompts to guide the base PLM, and incorporate KGs to improve the quality of prompts with a pre-training task.
Such a way can effectively endow the PLM with the background knowledge for better performance on the recommendation task.
Besides, we also use the response template generated by the conversation module as part of the prompt, which further improves the recommendation performance.
Note that our approach only tunes a few parameters compared with full parameter fine-tuning, hence it is also much more efficient than those PLM-based methods.

\paratitle{Ablation Study.}
Our approach designs a set of prompt components to improve the performance of CRS.
To verify the effectiveness of each component, we conduct the ablation study on the \textsc{ReDial} dataset, and report the results of Recall@10 and Recall@50.
We consider removing the pre-training task of the semantic fusion module, token or entity information in the fused knowledge representations, task-specific soft tokens, and the response template, respectively.

The results are shown in Figure~\ref{fig:ablation-rec}.
We can see that removing any component would lead to performance degradation.
It indicates that all the components in our model are useful to improve the performance of the recommendation task.
Among them, the performance decreases the most after removing the pre-training task in the semantic fusion module. It indicates that such a pre-training process is important in our approach, since it can learn the semantic correlations between entities and tokens, which enforces the entity semantics to be aligned with the base PLM.

\subsection{Evaluation on Conversation Task}
In this part, we conduct experiments to verify the effectiveness of our model on the conversation task.

\begin{table}[t]
    \centering
    \caption{
        Automatic evaluation results on the conversation task.
        We abbreviate Distinct-2,3,4 as Dist-2,3,4.
        Numbers marked with * indicate that the improvement is statistically significant compared with the best baseline (t-test with p-value < 0.05).
    }
    \label{tab:conv-table}
    \resizebox{\linewidth}{!}{%
        \begin{tabular}{lcccccc}
            \toprule
            Datasets & \multicolumn{3}{c}{ReDial} & \multicolumn{3}{c}{INSPIRED}                                                                         \\
            \midrule
            Models   & Dist-2                     & Dist-3                       & Dist-4          & Dist-2          & Dist-3          & Dist-4          \\
            \midrule
            ReDial   & 0.225                      & 0.236                        & 0.228           & 0.406           & 1.226           & 2.205           \\
            KBRD     & 0.281                      & 0.379                        & 0.439           & 0.567           & 2.017           & 3.621           \\
            KGSF     & 0.302                      & 0.433                        & 0.521           & 0.608           & 2.519           & 4.929           \\
            \midrule
            GPT-2    & 0.354                      & 0.486                        & 0.441           & 2.347           & 3.691           & 4.568           \\
            DialoGPT & 0.476                      & 0.559                        & 0.486           & 2.408           & 3.720           & 4.560           \\
            BART     & 0.376                      & 0.490                        & 0.435           & 2.381           & 2.964           & 3.041           \\
            \midrule
            UniCRS   & \textbf{0.492}*            & \textbf{0.648}*              & \textbf{0.832}* & \textbf{3.039}* & \textbf{4.657}* & \textbf{5.635}* \\
            \bottomrule
        \end{tabular}%
    }
\end{table}

\begin{table}[t]
    \centering
    \caption{Human evaluation results about the conversation task on the \textsc{ReDial} dataset. 
    Numbers marked with * indicate that the improvement is statistically significant compared with the best baseline (t-test with p-value < 0.05).}
    \label{tab:human-table}
        \begin{tabular}{lcc}
            \toprule
            \textbf{Models} & \textbf{Fluency} & \textbf{Informativeness} \\
            \midrule
            ReDial          & 1.31             & 0.98                     \\
            KBRD            & 1.21             & 1.16                     \\
            KGSF            & 1.49             & 1.39                     \\
            \midrule
            GPT-2           & 1.62             & 1.48                     \\
            DialoGPT        & 1.68             & 1.56                     \\
            BART            & 1.63             & 1.43                     \\
            \midrule
            UniCRS          & \textbf{1.72}$^{*}$             & \textbf{1.64}$^{*}$                     \\
            \bottomrule
        \end{tabular}%
\end{table}

\begin{figure}[t]
    \centering
    \includegraphics[width=0.49\linewidth]{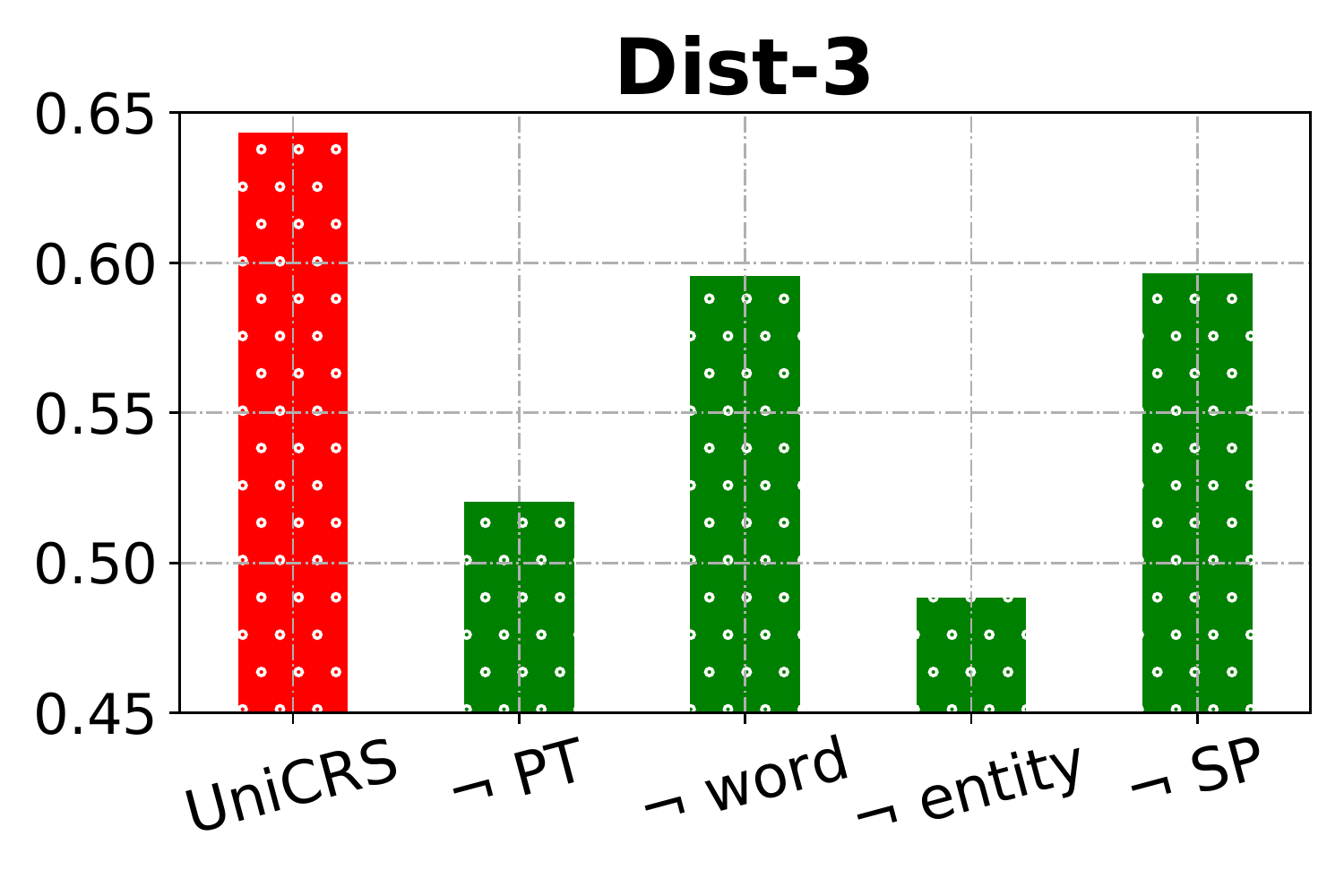}
    \includegraphics[width=0.49\linewidth]{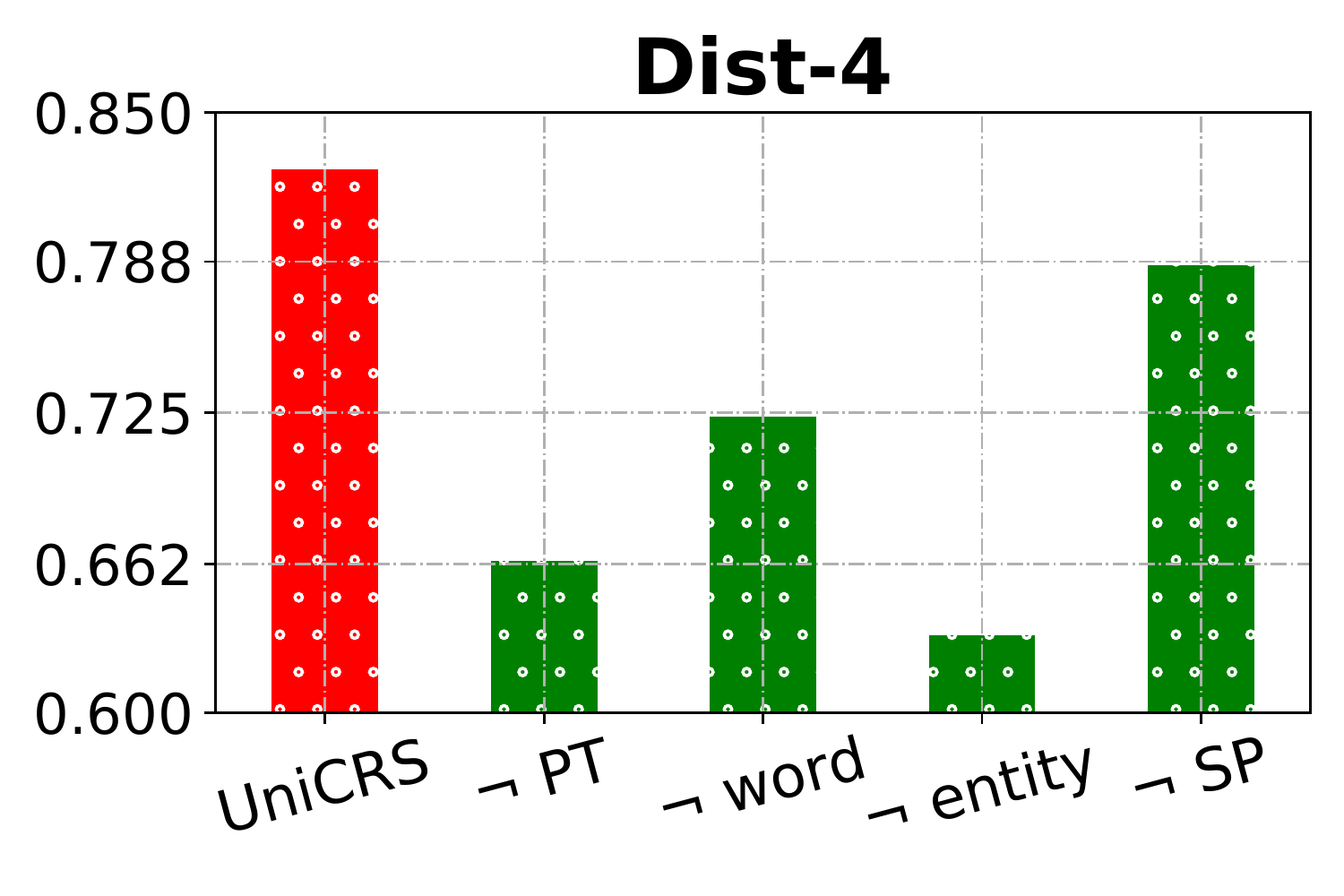}
    \caption{
        Ablation study on the \textsc{ReDial} dataset about the conversation task.
        PT denotes the pre-training task of sematic fusion.
        Word and entity refer to two kinds of data signals in the fusion module. 
        SP refers to task-specific soft tokens.
    }
    \label{fig:ablation-conv}
\end{figure}

\paratitle{Automatic Evaluation.}
We show the evaluation results of automatic metrics about different methods in Table~\ref{tab:conv-table}.
As we can see, among the three CRS methods, the performance order is also consistent with \emph{KGSF > KBRD > ReDial}. It is because KBRD adopts KG-based token bias to promote the probabilities of low-frequency tokens, and KGSF devises KG-enhanced cross-attention layers to improve the feature interactions of entities and tokens in the generation process.
Besides, we can see that PLMs achieve better performance than the three CRS methods. The possible reason is that they have been pre-trained with generative tasks on a large-scale general corpus, so they can quickly adapt to the CRS task and generate diverse responses after fine-tuning.
Among these PLMs, DialoGPT achieves the best performance. Since DialoGPT has been continually pre-trained on a large-scale dialogue corpus, it is more capable of generating informative responses in the CRS scenario.

Finally, compared with these baselines, our model also consistently performs better.
In our approach, we perform semantic fusion and prompt pre-training.
In this way, we can effectively inject task-specific knowledge into the PLM, and help generate informative responses.
Besides, since we only tune a few parameters compared with full parameter fine-tuning, we can alleviate the catastrophe forgetting problem of the PLM.

\paratitle{Human Evaluation.}
To further verify the effectiveness of our method, we conduct the human evaluation following previous works~\cite{zhou2020improving}. Table~\ref{tab:human-table} presents the results of human evaluation for the conversation task on the \textsc{ReDial} dataset.

First, among the three CRS methods, KGSF performs the best in both metrics, since it utilizes a KG-enhanced Transformer decoder that performs cross attention between the entity and word representations.
Besides, among the three PLM models, we can see that DialoGPT achieves the best performance. A possible reason is that DialoGPT has been continually pre-trained on a large-scale dialogue corpus, which endows it with a better capacity to generate high-quality responses.
Finally, our approach also outperforms all the baseline models. In our approach, we perform semantic fusion to inject the task-specific knowledge into DialoGPT, and also design a pre-training strategy to further enhance the prompt.
In this way, our model can effectively understand the dialogue history, and generate fluent and informative responses.

\paratitle{Ablation Study.}
In our approach, our proposed prompt design can also improve the performance of the conversation task.
To verify the effectiveness of each component, we conduct the ablation study on the \textsc{ReDial} dataset to analyze the contribution of each part.
We adopt Distinct-3 and Distinct-4 as the evaluation metrics, and consider removing the pre-training task of the semantic fusion module, token or entity information in the fused knowledge representations, and task-specific soft tokens, respectively.

The ablation results are shown in Figure~\ref{fig:ablation-conv}. 
We can see that removing any component would lead to a decrease in the model performance. It shows the effectiveness of all these components in our approach. 
Besides, the entity information seems to be more important than others, which yields a larger performance drop after being removed.  These entities contain domain-specific knowledge about items, which is helpful for our model to generate more informative responses.

\subsection{Performance Comparison w.r.t. Different Amount of Training Data}

\begin{figure}[t]
    \centering
    \includegraphics[width=0.49\linewidth]{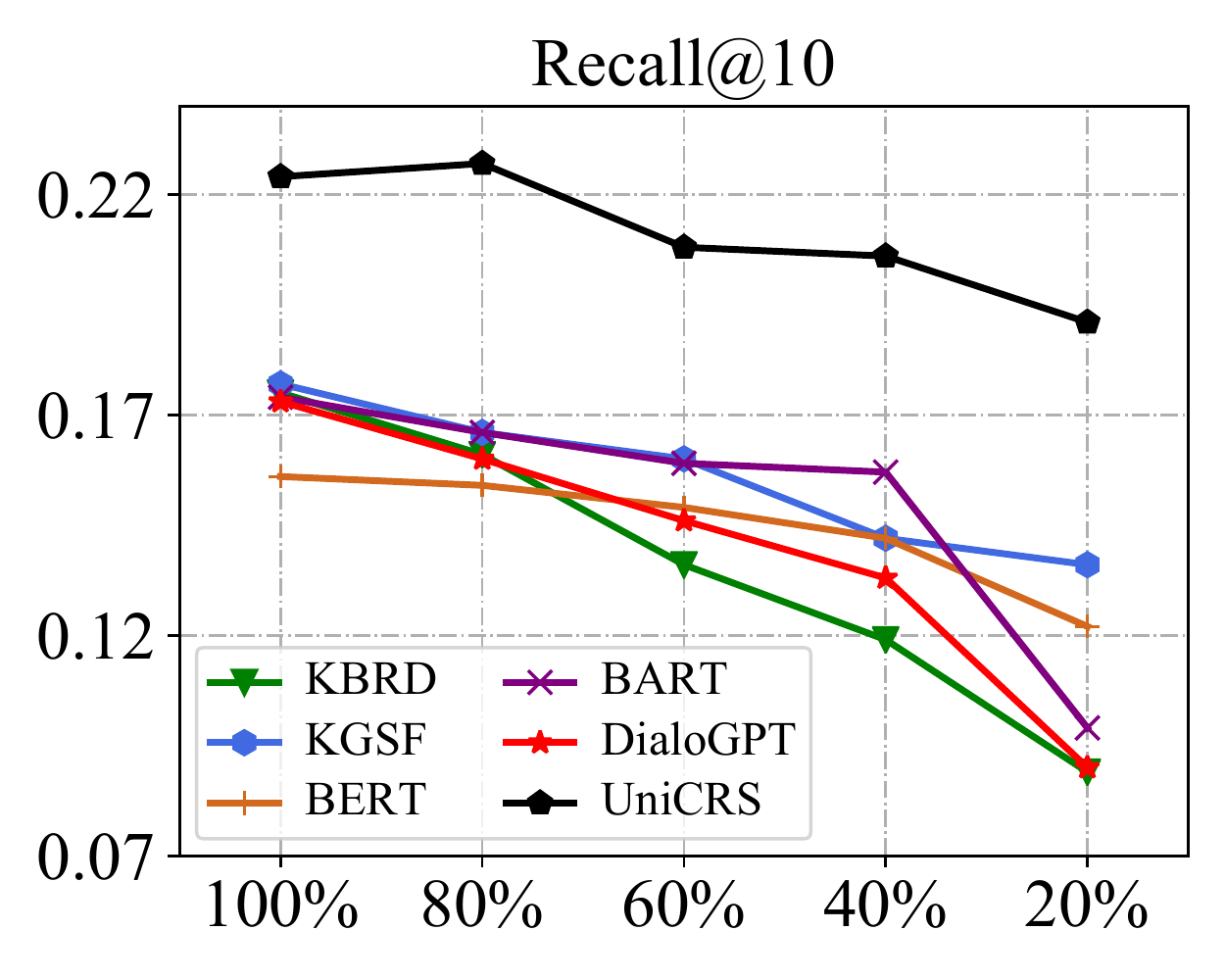}
    \includegraphics[width=0.49\linewidth]{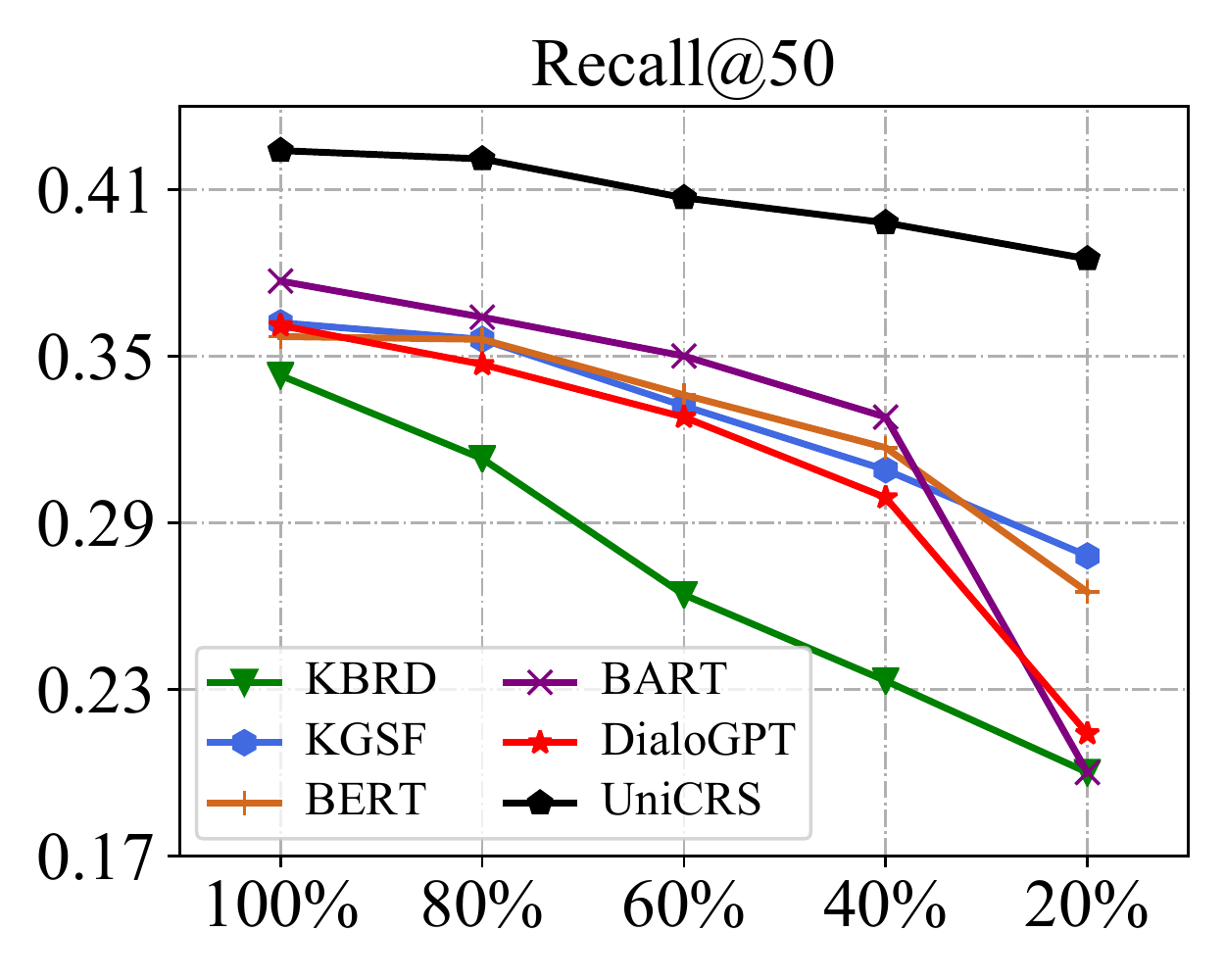}
    \caption{Performance comparison w.r.t. different amount of training data on \textsc{ReDial} dataset.}
    \label{fig:few-shot}
\end{figure}

Learning the parameters of CRSs requires a considerable amount of training data.
However, in real-world applications, it is likely to suffer from the cold start issue caused by insufficient data, which may increase the risk of overfitting.
Fortunately, since our approach only needs to optimize a few parameters in the prompt and incorporates a prompt pre-training strategy, the risk of overfitting can be reduced to some extent.
To validate this, we simulate a data scarcity scenario by sampling different proportions of the training data, and report the results of Recall@10 and Recall@50 on the \textsc{ReDial} dataset.

Figure~\ref{fig:few-shot} shows the evaluation results in different data scarcity settings.
As we can see, the performance of baseline models substantially drops with less available training data, while our method is consistently better than all the baseline models in all cases.
It indicates that our model can efficiently utilize the limited data and alleviate the cold start problem.
With extremely limited data (\ie 20\%), we find that our model still achieves a comparable performance with the best baseline that is trained with full data. It further indicates the effectiveness of our model in the cold start scenario.

\section{Conclusion}
In this paper, we proposed a novel conversational recommendation model named \textbf{UniCRS} to fulfill both the recommendation and conversation subtasks in a unified approach.  
First, taking a fixed PLM (\ie DialoGPT) as the backbone, we utilized a knowledge-enhanced prompt learning paradigm to reformulate the two subtasks.
Then, we designed multiple effective prompts to support both subtasks, which include fused knowledge representations generated by a pre-trained semantic fusion module, task-specific soft tokens, and the dialogue context.
We also leveraged the generated response template from the conversation subtask as an important part of the prompt to enhance the recommendation subtask.
The above prompt design can provide sufficient information about the dialogue context, task instructions, and background knowledge.
By only optimizing these prompts, our model can effectively accomplish both the recommendation and conversation subtasks.
Extensive experimental results have shown that our approach outperforms several competitive CRS and PLM methods, especially when only limited training data is available.

In the future, we will apply our model to more complicated scenarios, such as topic-guided CRS~\cite{zhou2020towards} and multi-modal CRS~\cite{yu2020towards}.
We will also consider devising more effective prompt pre-training strategies for quick adaptation to various CRS scenarios.

\section*{Acknowledgement}
This work was partially supported by Beijing Natural Science Foundation under Grant No. 4222027,  National Natural Science Foundation of China under Grant No. 61872369, and Beijing Outstanding Young Scientist Program under Grant No. BJJWZYJH012019100020098.
This work is also partially supported by Beijing Academy of Artificial Intelligence(BAAI).
Xin Zhao is the corresponding author.

\bibliographystyle{ACM-Reference-Format}
\bibliography{ref}


\end{document}